%% file: template-A4.tex
\documentclass[conference]{IEEEtran}
\IEEEoverridecommandlockouts
\usepackage{cite}
\usepackage{amsmath,amssymb,amsfonts}
\usepackage{algorithmic}
\usepackage{graphicx}
\usepackage{textcomp}
\usepackage{xcolor}
\usepackage[a4paper, total={184mm,239mm}]{geometry}

\usepackage{amsmath}

\usepackage{mathtools}
\usepackage{microtype}
\usepackage{booktabs}
\usepackage{enumitem}
\usepackage{tabularx}
\usepackage{hyperref}

\usepackage[ruled,vlined]{algorithm2e} 

\newcommand{\DecoHD}{\textsc{DecoHD}}

\def\BibTeX{{\rm B\kern-.05em{\sc i\kern-.025em b}\kern-.08em
    T\kern-.1667em\lower.7ex\hbox{E}\kern-.125emX}}

\makeatletter
\newcommand{\linebreakand}{%
  \end{@IEEEauthorhalign}
  \hfill\mbox{}\par
  \mbox{}\hfill\begin{@IEEEauthorhalign}
}

\begin{document}

\title{\textsc{DecoHD}: Decomposed Hyperdimensional Classification under Extreme Memory Budgets}

\author{\IEEEauthorblockN{Sanggeon Yun}
\IEEEauthorblockA{\textit{Dept. of Computer Science} \\
\textit{University of California, Irvine}\\
Irvine, CA, USA \\
sanggeoy@uci.edu}
\and
\IEEEauthorblockN{Hyunwoo Oh}
\IEEEauthorblockA{\textit{Dept. of Computer Science} \\
\textit{University of California, Irvine}\\
Irvine, CA, USA \\
hyunwooo@uci.edu}
\and
\IEEEauthorblockN{Ryozo Masukawa}
\IEEEauthorblockA{\textit{Dept. of Computer Science} \\
\textit{University of California, Irvine}\\
Irvine, CA, USA \\
rmasukaw@uci.edu}
\and
\IEEEauthorblockN{Mohsen Imani}
\IEEEauthorblockA{\textit{Dept. of Computer Science} \\
\textit{University of California, Irvine}\\
Irvine, CA, USA \\
m.imani@uci.edu}
}

\maketitle

\begin{abstract}
    \input{Sections/0_Abstract}
\end{abstract}


\section{Introduction}\label{sec:introduction}
\input{Sections/1_Introduction}

\section{Related Work}\label{sec:related}
\input{Sections/2_RelatedWork}

\begin{figure*}[t]
    \centering
    \includegraphics[width=0.8\linewidth]{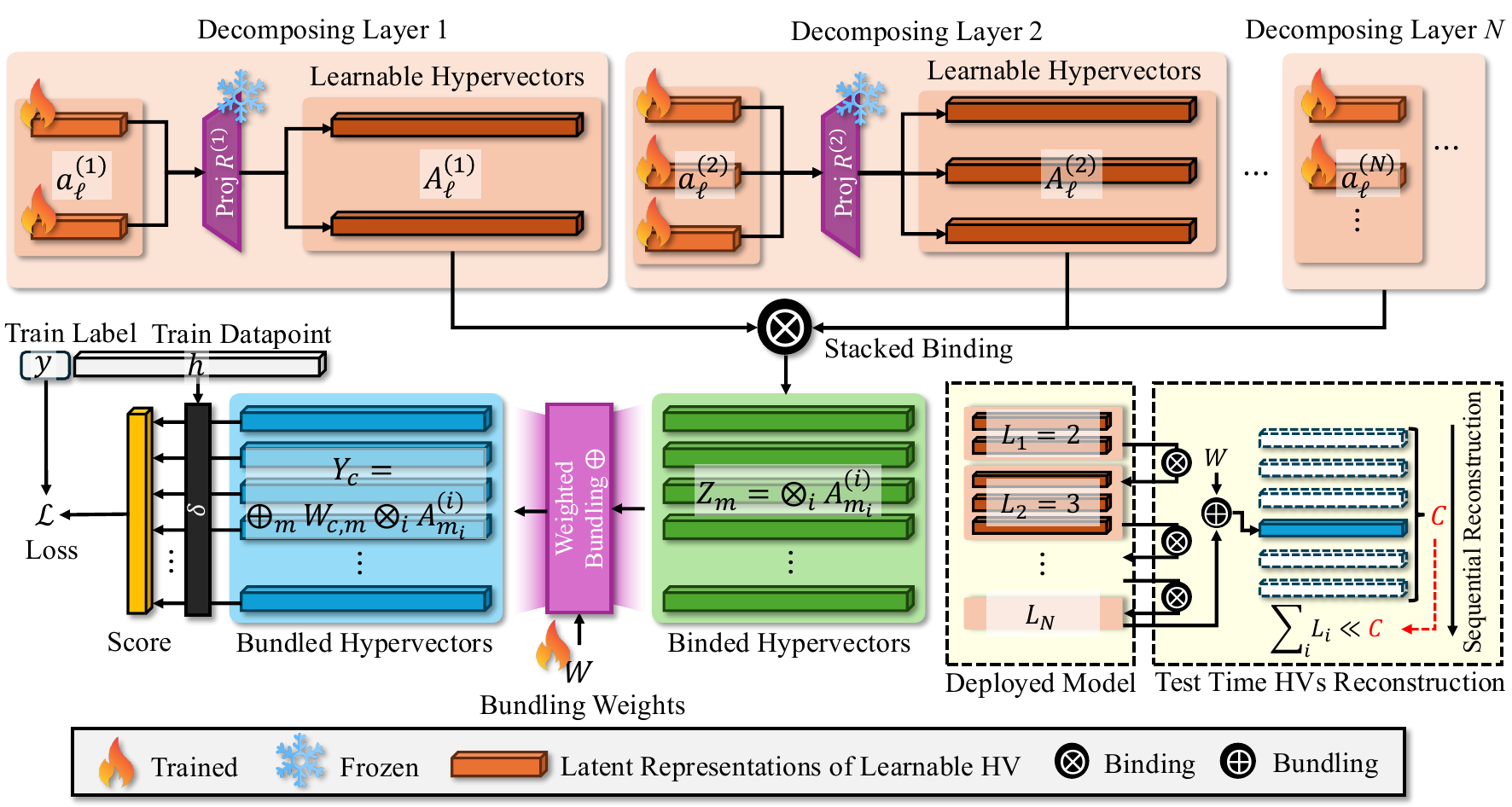}
    \caption{\textbf{\DecoHD{} overview.}
    A fixed encoder maps inputs to hypervectors $h\!\in\!\mathbb{R}^{D}$.
    $N$ layers each provide $L_i$ learnable channels $\{A^{(i)}_{\ell}\}$ generated from low-dimensional latents via frozen projections.
    For an input, all $M=\prod_i L_i$ path hypervectors are formed by successive binding ($\otimes$); class-wise vectors are then produced by weighted bundling ($\oplus$) with $W\!\in\!\mathbb{R}^{C\times M}$ and scored against $h$ via dot products.
    Training updates only the latents and $W$.}
    \label{fig:overview}
\end{figure*}

\section{Methodology}
\input{Sections/3_Methodology}

\section{Experiments}\label{sec:experiments}
\input{Sections/4_Experiments.tex}

\section{Conclusions}\label{sec:conclusions}
\input{Sections/5_Conclusions}

\section*{Acknowledgements}
This work was supported in part by the DARPA Young Faculty Award, the National Science Foundation (NSF) under Grants \#2431561,  \#2127780, \#2319198, \#2321840, \#2312517, and \#2235472, the Semiconductor Research Corporation (SRC), the Office of Naval Research through the Young Investigator Program Award and Grants \#N00014-21-1-2225 and \#N00014-22-1-2067, Army Research Office Grant \#W911NF2410360, the U.S. Army Combat Capabilities Development Command Army Research Laboratory under Support Agreement No. USMA 21050, and DARPA under Support Agreement No. USMA 23004. Additionally, support was provided by the Air Force Office of Scientific Research under Award \#FA9550-22-1-0253, along with generous gifts from Xilinx and Cisco.

\bibliographystyle{ieeetr}
\bibliography{mybibliography}

\end{document}

%% file: Sections/0_Abstract.tex
Decomposition is a proven way to shrink deep networks without changing input-output dimensionality or interface semantics. We bring this idea to hyperdimensional computing (HDC), where footprint cuts usually shrink the feature axis and erode concentration and robustness. Prior HDC decompositions decode via fixed atomic hypervectors, which are ill-suited for compressing learned class prototypes. We introduce \DecoHD{}, which learns directly in a decomposed HDC parameterization: a small, shared set of per-layer channels with multiplicative binding across layers and bundling at the end, yielding a large representational space from compact factors. \DecoHD{} compresses along the class axis via a lightweight bundling head while preserving native bind–bundle–score; training is end-to-end, and inference remains pure HDC, aligning with in/near-memory accelerators. In evaluation, \DecoHD{} attains extreme memory savings with only minor accuracy degradation under tight deployment budgets. On average it stays within about $0.1$–$0.15\%$ of a strong non-reduced HDC baseline (worst case $5.7\%$), is more robust to random bit-flip noise, reaches its accuracy plateau with up to $\sim97\%$ fewer trainable parameters, and—in hardware—delivers roughly $277\times/35\times$ energy/speed gains over a CPU (AMD Ryzen 9 9950X), $13.5\times/3.7\times$ over a GPU (NVIDIA RTX 4090), and $2.0\times/2.4\times$ over a baseline HDC ASIC.

%% file: Sections/1_Introduction.tex
Hyperdimensional computing (HDC), rooted in vector-symbolic architectures (VSA), has gained traction as a lightweight and noise-tolerant paradigm for classification and learning on constrained platforms~\cite{kanerva2009hyperdimensional,imani2019framework,hernandez2021onlinehd,yun2024neurohash,yun2025hyperdimensional,yun2024hypersense,yun2025missionhd,wang2023disthd}. By encoding data into dense high-dimensional hypervectors and manipulating them with simple algebraic operations, HDC classifiers combine attractive traits: inherent robustness to device-level non-idealities, parallelism amenable to hardware acceleration, and compact compute/memory footprints. These characteristics align naturally with the demands of energy-constrained or near-memory execution environments, where classical deep learning often proves too heavy~\cite{ahmed2024internet,custode2024fast,shi2024introduction,laskaridis2024future,masum2025parahdc}.

\begin{figure}[t]
  \centering
  \includegraphics[width=\linewidth]{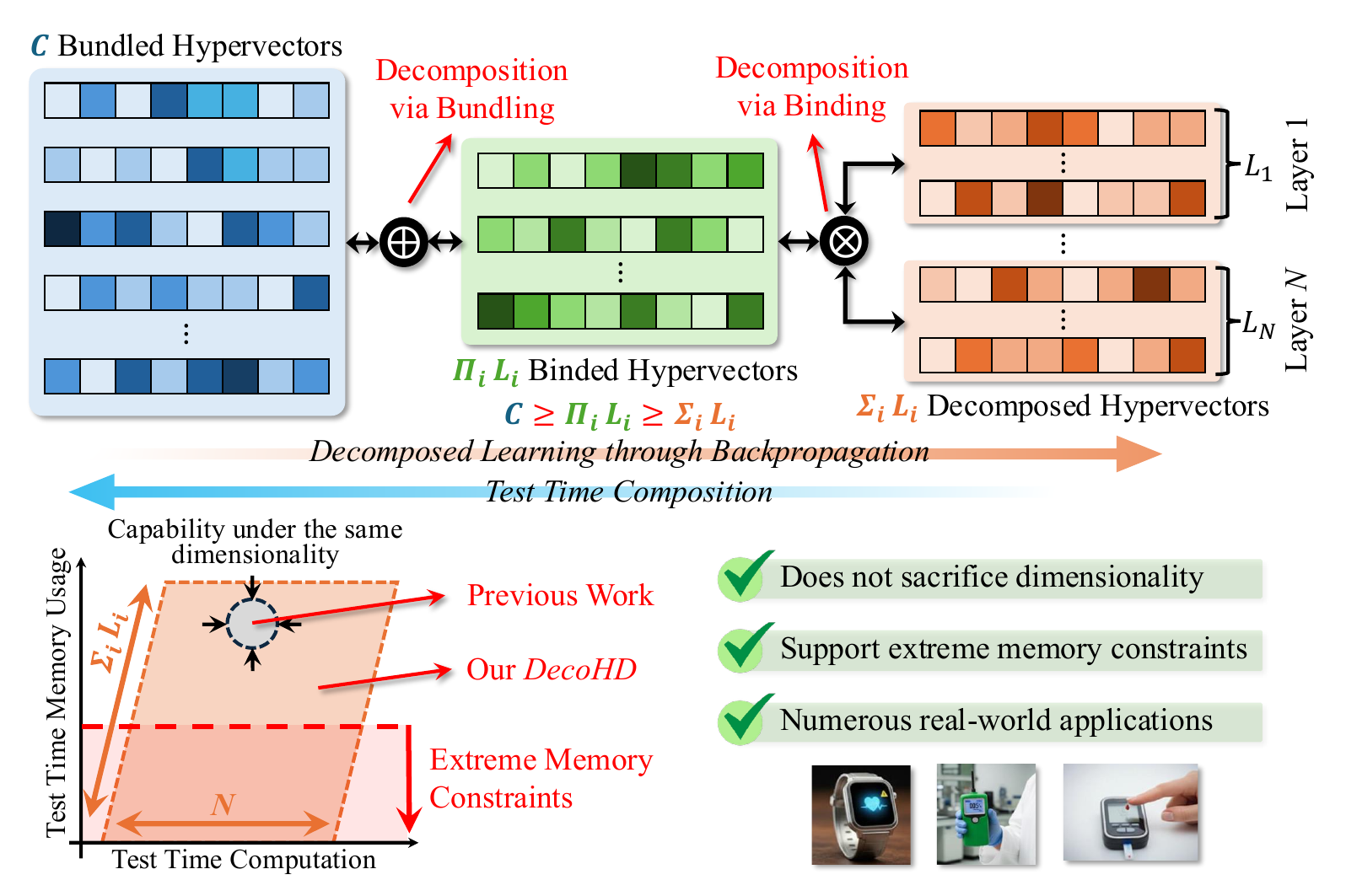}
  \caption{\textbf{Our proposed prototype reduction strategy.} Conventional models store one dense hypervector per class ($\mathcal{O}(CD)$). \DecoHD{} composes prototypes from a small shared set of channels, yielding $M=\prod_i L_i$ \emph{bound paths} with $\sum_i L_i \le M \le C$, reducing memory while preserving full-$D$ representations. As illustrated in the bottom-left figure, varying the number of layers $N$ and the channel count $\sum_i L_i$ enables a tunable trade-off among memory footprint, inference cost, accuracy, and robustness, whereas prior feature-axis methods collapse to a single point at fixed dimensionality.}
  \label{fig:comparison}
\end{figure}

A conventional HDC classifier stores one prototype hypervector per class and predicts by measuring similarity between a query and these prototypes, typically via dot product~\cite{hernandez2021onlinehd}. While this layout is simple and hardware-friendly, its storage requirement grows as $\mathcal{O}(CD)$ for $C$ classes and dimensionality $D$, which quickly dominates memory in multi-class problems. Prior work has largely reduced this footprint by compressing along the \emph{feature axis}: lowering $D$, imposing sparsity, or quantizing encodings~\cite{imani2019sparsehd,morris2019comphd,ge2020classification,verges2025classification,imani2019quanthd}. Although effective in reducing memory and multiply–accumulate cost, such strategies erode the concentration-of-measure properties that stabilize similarity in high dimensions. At extreme budgets, they degrade both robustness to analog noise~\cite{liang2024robust,liang2024stridehd} and classification accuracy.

This tension is especially acute in deployment scenarios where memory is the first-order constraint. Across emerging AI-on-edge applications—such as TinyML-based sensing and recognition~\cite{celen2025tinymltraffic,li2025tinymlpositioning,liu2025solarml}, batteryless IoT platforms~\cite{custode2024fast,tabrizchi2025intermittent}, and federated/distributed AI in heterogeneous systems~\cite{dudziak2022fedoras,zhu2024shufflefl,chu2025priprune}—models must be aggressively compressed while retaining resilience to device-level imperfections. In these regimes, preserving $D$ maintains the concentration-of-measure properties central to HDC, making class-axis compression a more principled pathway than shrinking dimensionality.

Inspired by decomposition techniques successful in compact deep neural networks—such as low-rank adaptation, modular fine-tuning, and shared-codebook representations~\cite{yin2021towards,yin2022batude,horvath2023maestro,tastan2025loft,kwon2023tinytrain}—we propose a complementary direction tailored to HDC. Rather than compressing hypervector dimensionality, we reduce the \emph{number} of stored dense hypervectors by parameterizing class prototypes as structured compositions of a small, shared set of \emph{channels}, combined using HDC-native binding ($\otimes$) and weighted bundling ($\oplus$), as illustrated in \autoref{fig:comparison}. Concretely, $N$ layers expose $\{L_i\}$ learnable channels; selecting one channel per layer yields $M=\prod_i L_i$ \emph{bound paths}. A lightweight class-specific head $W \in \mathbb{R}^{C\times M}$ aggregates these paths to form logits. This design scales memory roughly with $\sum_i L_i$, while maintaining full-$D$ representations and their robustness benefits. This decomposition also introduces a tunable memory–compute trade-off: increasing the number of layers $N$ or channels reduces storage but raises test-time computation, since more bound-path compositions must be evaluated; deeper factorizations with small $L$ provide larger representational capacity at the cost of higher inference cost, a favorable exchange in regimes where memory is the primary constraint.

Naively decomposing dense prototypes post hoc is unstable in high dimensions, since binding is multiplicative and orderless, making the inverse factorization ill-posed~\cite{frady2020resonator,poduval2024hdqmf}. Instead, we introduce \DecoHD{}, which learns \emph{directly in the decomposed space}. Low-dimensional latents are expanded into channel hypervectors via fixed random projections, and training optimizes the bind–bundle pipeline end-to-end using cross-entropy. At inference, the method relies solely on $\otimes$, $\oplus$, and dot products, ensuring compatibility with existing HDC/VSA accelerators and memory-centric hardware paths.

In summary, \DecoHD{} tackles the memory bottleneck of conventional HDC by decomposing class prototypes into a compact set of shared channels while preserving the ambient dimensionality $D$. This design retains the stability and robustness of high-dimensional representations under extreme budgets and aligns with the needs of edge-AI deployments across TinyML, federated learning, and memory-centric accelerators, where compact yet reliable models are critical~\cite{laskaridis2024future}. Our evaluation shows that \DecoHD{} achieves accuracy comparable to conventional HDC while delivering up to $277\times$ higher energy efficiency and $35\times$ faster inference than CPU baselines, $14\times$ and $3.7\times$ improvements over GPU, and $2.0\times$ and $2.4\times$ gains over a conventional HDC ASIC, all with only $0.38\times$ memory.

The main contributions of this work are:
\begin{enumerate}[leftmargin=*,itemsep=2pt,topsep=2pt]
    \item \textbf{Decomposed HDC classifier.} We propose \DecoHD{}, which replaces $C$ dense prototypes with a compact set of shared \emph{channels} and a lightweight bundling head. This reduces memory from $\mathcal{O}(CD)$ to $\mathcal{O}(LD)$ with $L\!\ll\!C$, while preserving dimensionality $D$ and robustness to device-level noise.
    
    \item \textbf{End-to-end HDC-aware training.} We develop a bind–bundle pipeline that jointly optimizes low-dimensional latents (expanded via fixed random projections) and class bundling weights. Unlike post hoc factorization, this preserves holographic properties of HDC and enables stable learning directly in decomposed space, reducing trainable parameters by up to 97\%.
    
\end{enumerate}

%% file: Sections/2_RelatedWork.tex
\subsection{HDC and Vector Symbolic Architectures}
Hyperdimensional computing—also known as vector symbolic architectures—represents symbols and compositional structure with high-dimensional hypervectors manipulated by \emph{binding} ($\otimes$), \emph{bundling} ($\oplus$), and permutation ($\pi$)~\cite{kanerva2009hyperdimensional}.
Classification typically superposes encoded examples into per-class prototypes and scores by dot product.
This operation set maps well to emerging in/near-memory platforms and low-power edge contexts: recent work demonstrates robust HDC pipelines under voltage scaling and binary encodings~\cite{liang2024stridehd}, specialized VSA/HDC designs for efficiency~\cite{moghadam2025idvsa}, and accelerator paths that preserve the native HDC interface~\cite{masum2025parahdc}.
At the system level, the push toward consumer edge-AI~\cite{laskaridis2024future} and intermittently powered/batteryless devices~\cite{ahmed2024internet,custode2024fast,tabrizchi2025intermittent} further motivates lightweight, memory-frugal models.
Concurrently, the embedded-systems community highlights in/near-memory directions~\cite{shi2024introduction} and 3D integration trends~\cite{yin2021towards,yin2022batude} that favor linear, data-parallel HDC kernels.
Our design intentionally keeps HDC’s native bind–bundle–dot scoring to remain drop-in compatible with such accelerators.

\subsection{Model-Size Reduction for HDC-based Classifiers}
Many efficiency efforts in HDC implicitly shrink along the \emph{feature axis}—e.g., binary/low-precision encodings or simplified encoders—which reduce stored bits and arithmetic but also reduce effective dimensionality, weakening high-$D$ averaging that stabilizes similarity and increasing sensitivity to perturbations under tight power/voltage budgets~\cite{liang2024stridehd,zhang2021assessing}.
In contrast, \DecoHD{} preserves the ambient $D$ and compresses orthogonally along the \emph{class axis} by sharing a small pool of per-layer channels across classes and learning only the light bundling head.
This reduces the \# of dense hypervectors without altering the bind–bundle–score pipeline, and complements systems work on efficient HDC datapaths and accelerators~\cite{masum2025parahdc,moghadam2025idvsa}.

\subsection{Decomposition in DNNs and HDC for Model Compression}
Decomposition is a standard route to shrink DNNs while preserving input/output dimensionality—e.g., trainable low-rank/tensor factorizations and related adapters~\cite{horvath2023maestro,tastan2025loft}.
Our approach brings this spirit to HDC but avoids ill-posed post hoc factorization of dense class prototypes.
Instead, \DecoHD{} \emph{learns in a decomposed parameterization}: a small, shared set of per-layer channels whose stacked bindings yield $M$ path features, followed by class-wise weighted bundling.
This keeps $D$ unchanged, retains native HDC operations ($\otimes,\oplus$), and shifts storage from $C\times D$ prototypes to compact channels plus a light bundling head—well-suited to near-memory execution and resource-constrained edge deployment~\cite{shi2024introduction,laskaridis2024future,ahmed2024internet}.
Finally, our end-to-end training of low-dimensional latents (expanded by fixed random projections) aligns with recent trends in efficient/edge training and personalization~\cite{kwon2023tinytrain,dudziak2022fedoras,zhu2024shufflefl}.

%% file: Sections/3_Methodology.tex
\begin{algorithm}[t]
\small
\caption{\DecoHD{}: Training \& Inference}
\label{alg:decohd}
\DontPrintSemicolon
\SetKwInOut{Input}{Input}\SetKwInOut{Output}{Output}

\Input{Data $\mathcal{D}=\{(x,y)\}$, fixed encoder $\phi:\mathbb{R}^{d_{\text{in}}}\!\to\!\mathbb{R}^{D}$, layers $N$, channels $\{L_i\}_{i=1}^{N}$, latent dim $d$, fixed projectors $R^{(i)}\!\in\!\mathbb{R}^{d\times D}$.}
\Output{Latents $\{a^{(i)}_{\ell}\!\in\!\mathbb{R}^{d}\}$ and bundling weights $W\!\in\!\mathbb{R}^{C\times M}$ with $M=\prod_{i=1}^{N}L_i$.}

\SetKwBlock{Init}{Init:}{}
\Init{$a^{(i)}_{\ell}\!\sim\!\mathcal{N}(0,\sigma^2)$,\; $W_{c,m}\!\leftarrow\!1/M$;\; freeze $\phi$ and all $R^{(i)}$. \;}

\SetKwBlock{Train}{Training (for epochs):}{}
\Train{\begin{enumerate}[leftmargin=*,itemsep=1pt]
\item Sample mini-batch $\{(x_b,y_b)\}_{b=1}^{B}$; encode $h_b\!\leftarrow\!\phi(x_b)$.
\item Compose channels: $A^{(i)}_{\ell}\!\leftarrow\! a^{(i)}_{\ell} R^{(i)} \in \mathbb{R}^{D}$ for all $i,\ell$.
\item For each $b$: form all path HVs by stacked binding
      $Z_{m}(h_b) \leftarrow h_b \otimes \bigotimes_{i=1}^{N} A^{(i)}_{m_i}$ for $m\!=\!1..M$.
\item Bundle per class and score:
      $Y_c(h_b)\!\leftarrow\!\bigoplus_{m=1}^{M} W_{c,m} Z_m(h_b)$,\;
      $s_c(x_b)\!\leftarrow\!\langle Y_c(h_b),\,h_b\rangle$.
\item Loss: $\mathcal{L}\!=\!\frac{1}{B}\sum_b \!-\log \frac{e^{s_{y_b}(x_b)}}{\sum_{c} e^{s_c(x_b)}}$;\;
      update $\{a^{(i)}_{\ell}\}$ and $W$ with AdamW.
\end{enumerate}}

\SetKwBlock{Inference}{Inference (Predict$(x_q)$):}{}
\Inference{\begin{enumerate}[leftmargin=*,itemsep=1pt]
\item $h\!\leftarrow\!\phi(x_q)$;\; materialize $\{A^{(i)}_{\ell}\}$.
\item Compute $Z_m(h)$ sequantially;\; $Y_c(h)\!\leftarrow\!\bigoplus_{m} W_{c,m} Z_m(h)$;\;
      $s_c\!\leftarrow\!\langle Y_c(h), h\rangle$.
\item \Return $\hat{y}\!=\!\arg\max_{c} s_c$.
\end{enumerate}}
\end{algorithm}

Conventional HDC stores one $D$-dimensional prototype per class, yielding $\mathcal{O}(CD)$ memory.
Directly \emph{decomposing} trained class hypervectors into a small set of shared factors is ill-posed in HD spaces: binding is multiplicative and orderless, and many distinct factorizations yield near-identical similarities.
\DecoHD{} sidesteps this by learning \emph{directly in a decomposed parameterization} as illustrated in \autoref{fig:overview}.
We introduce $N$ layers; layer $i$ exposes $L_i$ learnable \emph{channels}—each channel is a factor hypervector that can bind with the input.
Selecting one channel per layer defines a \emph{path}; all $M=\prod_{i=1}^{N}L_i$ paths are formed by stacked binding and then \emph{bundled} with class-specific weights to produce class vectors.
This turns a small linear budget in $\sum_i L_i$ channels into a combinatorial set of $M$ bound paths while keeping the HD operations native. Detailed procedures for each stage is described in \autoref{alg:decohd}.

\subsection{Preliminaries}
Let $D$ denote the hypervector dimension and $C$ the number of classes.
HDC represents items as $D$-dimensional real hypervectors and uses two primitives:
\emph{binding} ($\otimes$) as elementwise multiplication and \emph{bundling} ($\oplus$) as (weighted) elementwise addition.
An encoder $\phi:\mathbb{R}^{d_{\mathrm{in}}}\!\rightarrow\!\mathbb{R}^{D}$ maps inputs to hypervectors; we use a \emph{fixed} random projection $\phi(x)=xW_{\mathrm{enc}}$ (Gaussian or ternary).
Similarity is measured by the \textbf{dot product}, and we train with \textbf{cross-entropy}.

We write $N$ for the number of layers, $L_i$ for the number of channels in layer $i$, and
\begin{equation}
M \;=\; \prod_{i=1}^{N} L_i
\end{equation}
for the number of binding paths.
The class-wise bundling weights are $W\!\in\!\mathbb{R}^{C\times M}$.

\subsection{\DecoHD{} Train}
\noindent\textbf{Channels from low-dimensional latents.}
Layer $i$ contains $L_i$ low-dimensional latents $a^{(i)}_{\ell}\!\in\!\mathbb{R}^{d}$ with $d\!\ll\! D$.
Each latent is expanded by a \emph{frozen} random projector $R^{(i)}\!\in\!\mathbb{R}^{d\times D}$ to a channel (factor) hypervector:
\begin{equation}
A^{(i)}_{\ell} \;=\; a^{(i)}_{\ell}\, R^{(i)} \;\in\; \mathbb{R}^{D}.
\label{eq:channel}
\end{equation}
Only the latents $\{a^{(i)}_{\ell}\}$ and the class bundling weights $W$ are learned; all $R^{(i)}$ and the input encoder $W_{\mathrm{enc}}$ remain fixed.

\medskip
\noindent\textbf{Path composition by stacked binding.}
Given $h=\phi(x)$, picking one channel per layer yields a path $m=(m_1,\dots,m_N)\in [L_1]\!\times\!\cdots\!\times\![L_N]$.
We construct the path hypervector by binding $h$ with the selected channels:
\begin{equation}
Z_m(h) \;=\; h \;\otimes\; \bigotimes_{i=1}^{N} A^{(i)}_{m_i} \;\in\; \mathbb{R}^{D}.
\label{eq:path}
\end{equation}
Broadcasted binding realizes all $M$ paths layer-by-layer.

\medskip
\noindent\textbf{Class bundling and logits.}
Class $c$ aggregates path hypervectors with learned weights $W_{c,m}$:
\begin{equation}
Y_c(h) \;=\; \bigoplus_{m=1}^{M} W_{c,m}\, Z_m(h) \;\in\; \mathbb{R}^{D},
\label{eq:bundle}
\end{equation}
and we compute dot-product logits
\begin{equation}
s_c(x) \;=\; \langle Y_c(h),\, h \rangle .
\label{eq:logit}
\end{equation}

\medskip
\noindent\textbf{Training.} During training, we minimize cross-entropy over logits \autoref{eq:logit}:
\begin{equation}
\mathcal{L}_{\mathrm{CE}}(x,y)
\;=\;
-\log \frac{\exp\big(s_{y}(x)\big)}{\sum_{c=1}^{C} \exp\big(s_c(x)\big)}.
\end{equation}
Gradients backpropagate through the $\otimes\!\rightarrow\!\oplus$ pipeline to the latents $\{a^{(i)}_{\ell}\}$ and $W$.
Keeping $R^{(i)}$ and $W_{\mathrm{enc}}$ fixed (i) preserves holography in the HD space and (ii) makes optimization efficient because the learnable parameters reside in the low-dimensional latent space.

\subsection{\DecoHD{} Inference}
At test time we materialize the channels $\{A^{(i)}_{\ell}\}$ once from the trained latents and perform \emph{sequential} test-time composition under extreme memory budgets. Rather than materializing all $M$ path features, we stream them one-by-one:

$$
\begin{aligned}
\underbrace{\text{stacked binding}}_{\otimes}
&\;\Rightarrow\;
\underbrace{\text{class bundling (accumulate)}}_{\oplus}
\\&\;\Rightarrow\;
\underbrace{\text{dot-product scoring}}_{\langle\cdot,\cdot\rangle}.
\end{aligned}
$$

Concretely, we iterate over $m=1,\dots,M$: (i) form $Z_m(h)$ via \autoref{eq:path} by sequentially binding the selected channels; (ii) immediately accumulate into class bundles $Y_c \!\leftarrow\! Y_c \oplus W_{c,m} Z_m(h)$ via \autoref{eq:bundle}; after the sweep, (iii) compute logits $s_c(h)=\langle Y_c(h),h\rangle$ (\autoref{eq:logit}) and return $\arg\max_c s_c(x)$. All operations stay in $\mathbb{R}^{D}$ and use native HDC primitives. 
We can further shrink peak memory by conducting \emph{score-only streaming}, skipping storing $Y_c$ and updating scores directly as $s_c \!\leftarrow\! s_c + W_{c,m}\,\langle Z_m(h),h\rangle$, discarding $Z_m(h)$ after each step.

\subsection{Memory and Compute}
A conventional HDC table stores $C$ prototypes, i.e., $\mathcal{O}(CD)$ reals. \DecoHD{} stores (i) $\sum_{i=1}^{N} L_i$ shared channels in $\mathbb{R}^{D}$ (or, during training, only low-dimensional latents plus fixed $R^{(i)}$), and (ii) a lightweight bundling head $W\!\in\!\mathbb{R}^{C\times M}$ with $M=\prod_i L_i$. 
Per sample, we use only native HDC primitives, but the arithmetic differs from a prototype table (which needs only $C$ dot products). In the streaming regime we iterate over paths $m=1,\dots,M$: (i) form $Z_m(h)$ by $N$ element-wise bindings, (ii) compute $t_m=\langle Z_m(h),h\rangle$, and (iii) update scores $s_c \leftarrow s_c + W_{c,m}\,t_m$. Streaming keeps peak memory to one working hypervector ($\mathcal{O}(D)$) plus $C$ scalars, trading extra per-sample work for a much smaller footprint than storing $C$ full prototypes. If memory permits pre-materializing class prototypes $Y_c$, inference reduces to the conventional $C$ dot products.

%% file: Sections/4_Experiments.tex
\begin{table}[t]
\centering
\caption{Datasets used in evaluations. $C$ denotes the number of classes.}
\label{tab:datasets}
\resizebox{\columnwidth}{!}{
\begin{tabular}{lrrrrl}
\toprule
\textbf{Dataset} & \textbf{\# Features} & $\boldsymbol{C}$ & \textbf{\# Train} & \textbf{\# Test} & \textbf{Description} \\
\midrule
ISOLET~\cite{isolet_54}  & 617 & 26 & 6{,}238   & 1{,}559   & Voice recognition \\
UCIHAR~\cite{human_activity_recognition_using_smartphones_240}  & 261 & 12 & 6{,}213   & 1{,}554   & Activity recognition (mobile) \\
PAMAP2~\cite{pamap2_physical_activity_monitoring_231}  & 75  & 5  & 611{,}142 & 101{,}582 & Activity recognition (IMU) \\
PAGE~\cite{page_blocks_classification_78}    & 10  & 5  & 4{,}925   & 548      & Page layout blocks classification \\
\bottomrule
\end{tabular}
}
\end{table}

\begin{figure}
    \centering
    \includegraphics[width=1.0\linewidth]{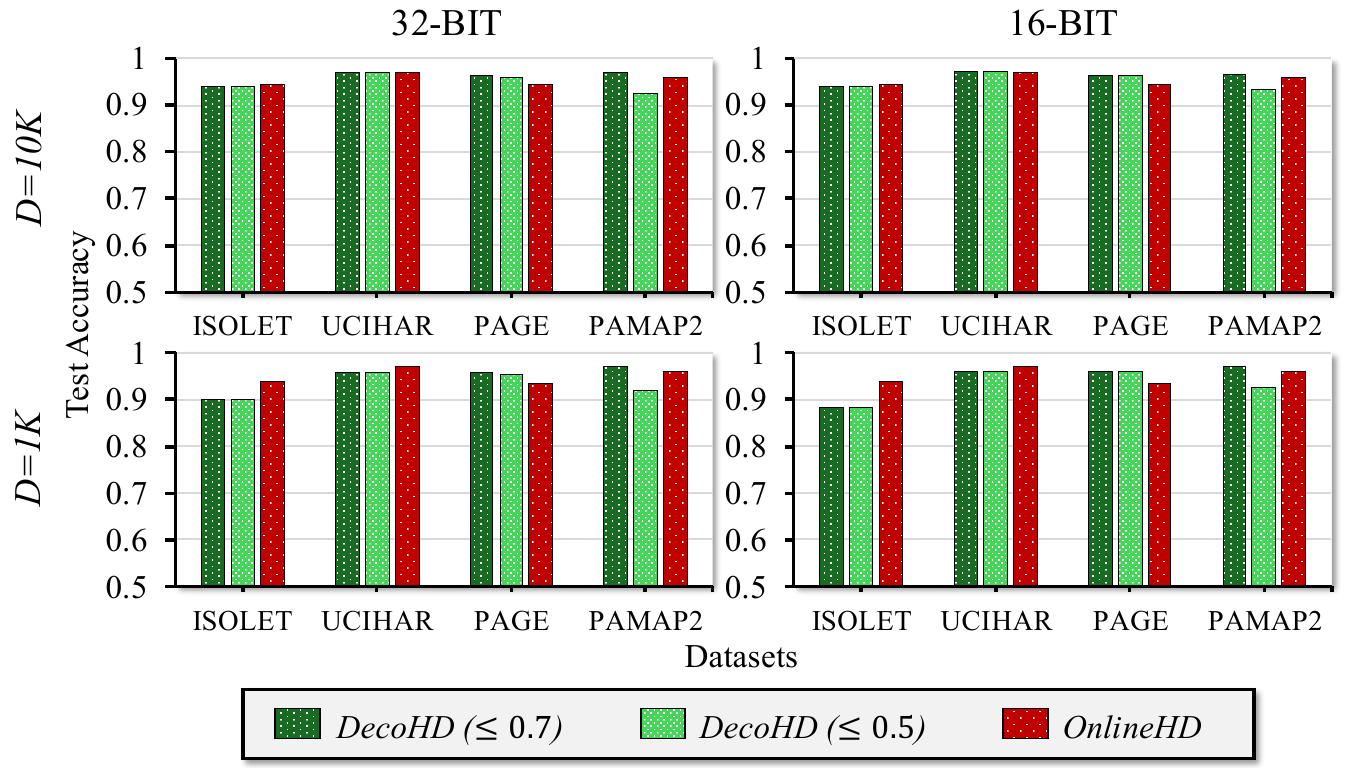}
    \caption{\textbf{Accuracy versus a strong HDC baseline.}
    Values in parentheses denote the target memory budget $m$ enforced relative to a conventional prototype table.
    We report test accuracy across numeric precisions and hypervector dimensionalities $D$, and compare to \emph{OnlineHD} (no parameter reduction).}
    \label{fig:decohd_perf}
\end{figure}

\begin{figure}
    \centering
    \includegraphics[width=1.0\linewidth]{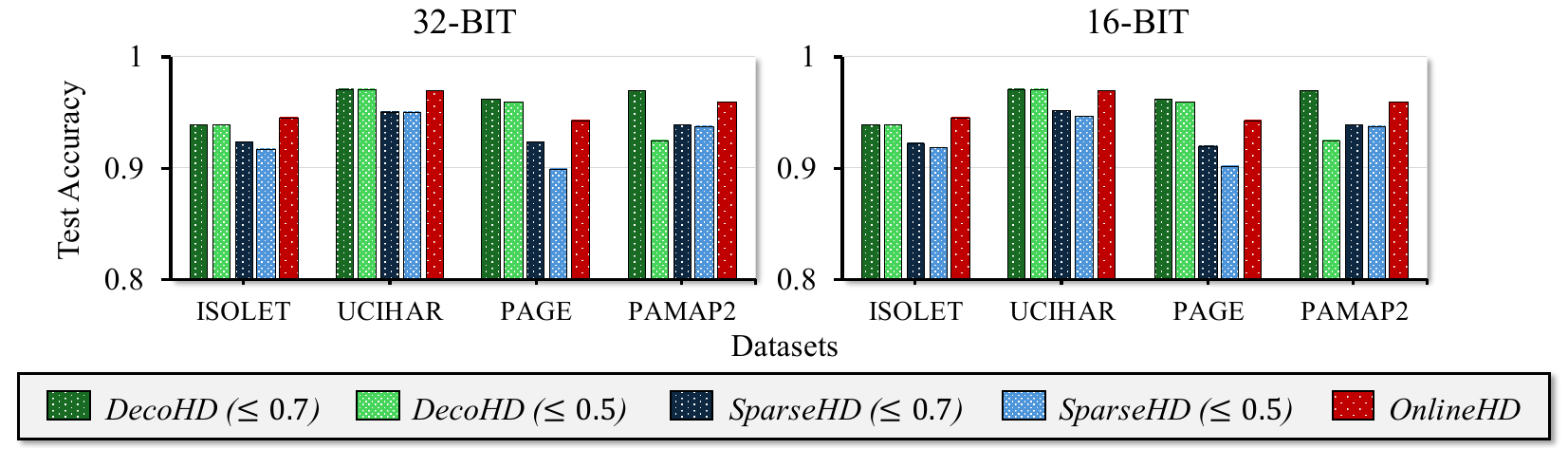}
    \caption{\textbf{Accuracy compared to state-of-the-art feature-axis compression.}
    Values in parentheses denote the target memory budget $m$.
    To isolate the impact of class-axis decomposition, we compare against \emph{SparseHD}, a representative feature-axis reduction method, under matched budgets.}
    \label{fig:decohd_perf_comp}
\end{figure}

\begin{figure}
    \centering
    \includegraphics[width=1.0\linewidth]{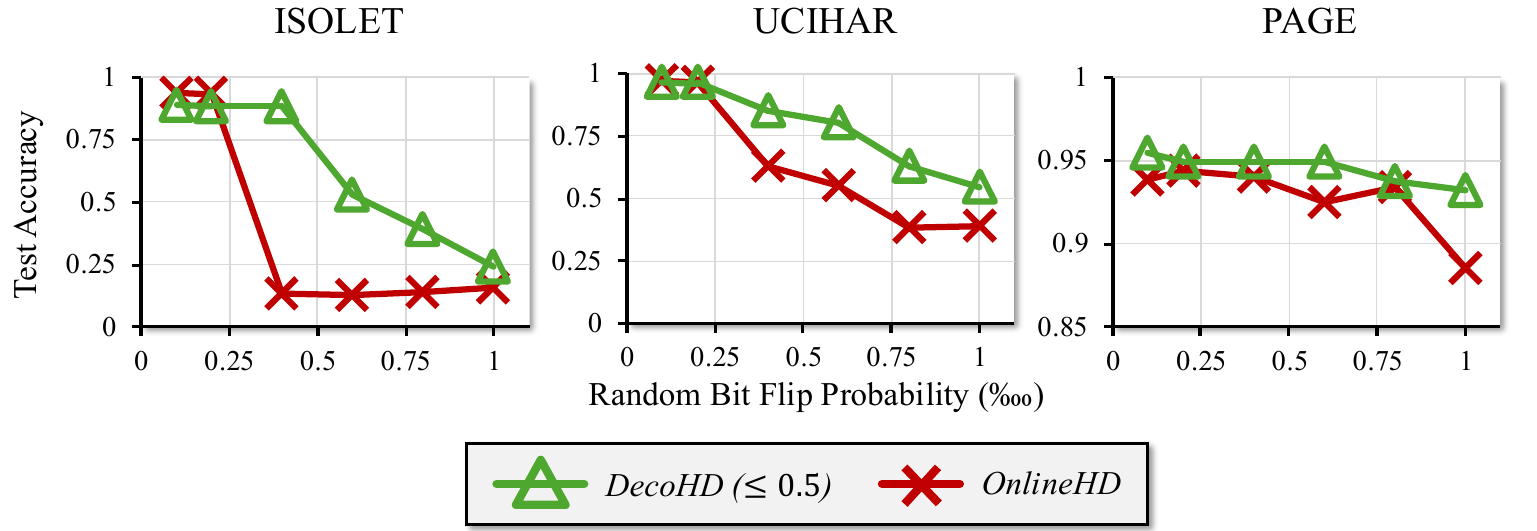}
    \caption{\textbf{Robustness to random bit-flip noise.}
    Values in parentheses denote the target memory budget $m$.
    We inject independent random bit flips into 32-bit floating-point representations and evaluate accuracy as the flip probability increases.}
    \label{fig:decohd_robust}
\end{figure}

\begin{figure}
    \centering
    \includegraphics[width=1.0\linewidth]{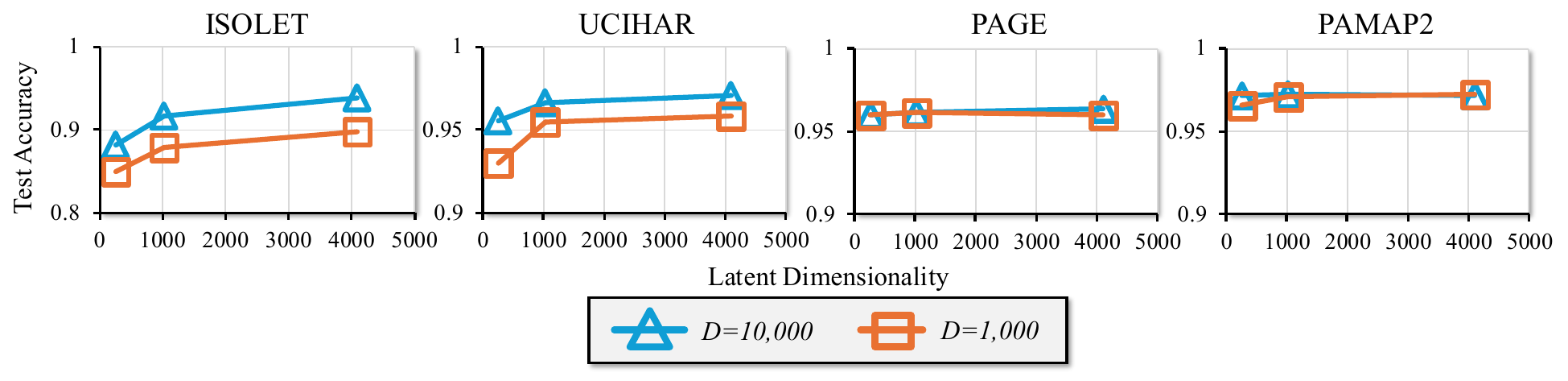}
    \caption{\textbf{Sensitivity to latent dimensionality.}
    We evaluate a two-layer $\lfloor\sqrt{C}\rfloor \times \lfloor\sqrt{C}\rfloor$ configuration with $d \in \{256, 1024, 4096\}$ at $D \in \{1{,}000, 10{,}000\}$ and report test accuracy.}
    \label{fig:decohd_ab_latent}
\end{figure}

\begin{figure}
    \centering
    \includegraphics[width=1.0\linewidth]{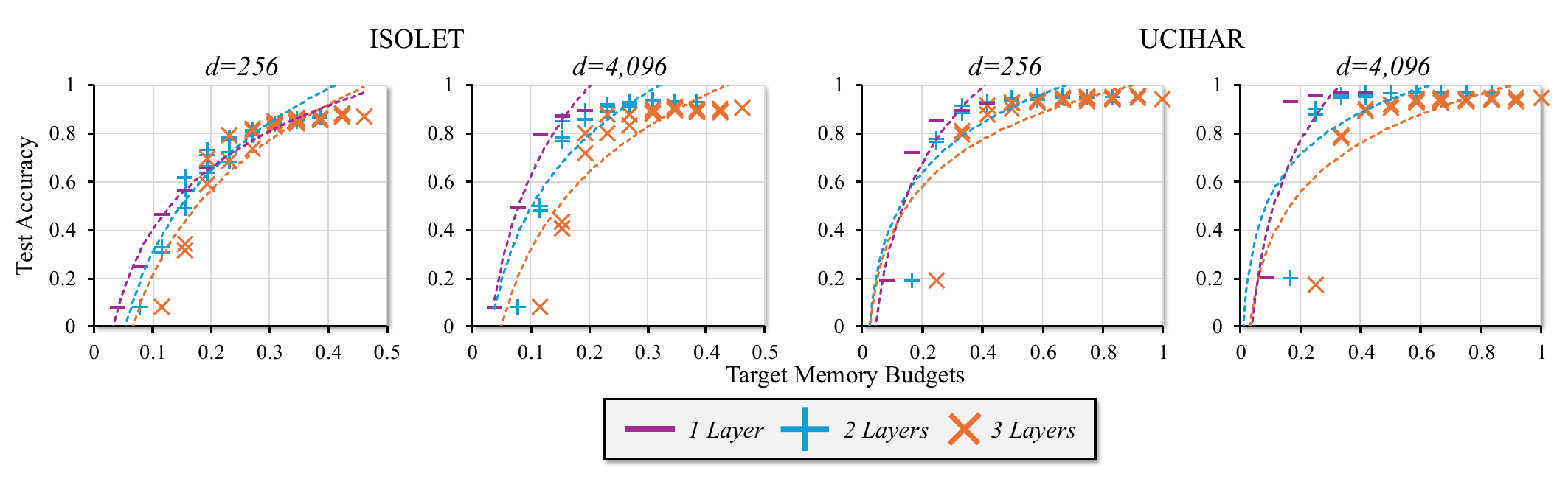}
    \caption{\textbf{Sensitivity to the number of layers.}
    We vary the number of layers ($N\in\{1,2,3\}$) under multiple target memory budgets, with up to five channels per layer, and report test accuracy for $d \in \{256, 4096\}$.
    The dotted curve is a logarithmic fit for visualization.}
    \label{fig:decohd_ab_layers}
\end{figure}

\subsection{Experimental Setup}

\medskip
\noindent\textbf{Implementation and precision.}
All models are implemented in \texttt{PyTorch} with a precision-aware training loop.
Unless otherwise stated, training uses 32-bit floating point with AdamW.
We additionally evaluate reduced-precision execution via native \texttt{fp16}/\texttt{bf16} (AMP on GPUs and \texttt{bf16} on CPU) and emulated \texttt{fp8}/\texttt{fp4} by rounding mantissa/exponent bits.
When available, we also include CUDA \texttt{fp8\_e4m3fn}/\texttt{fp8\_e5m2}.
No integer quantization or post-training calibration is applied.

\medskip
\noindent\textbf{Datasets and preprocessing.}
We follow the standard train/test splits for ISOLET, UCIHAR, PAGE, and PAMAP2 (see \autoref{tab:datasets} for details).
Features are standardized to zero mean and unit variance using statistics from the training split only, and all results are reported on the held-out test split.

\medskip
\noindent\textbf{Encoder.}
Inputs are mapped to $D$-dimensional real hypervectors using a \emph{fixed} random encoder $\phi(x){=}xW_{\text{enc}}$ (Gaussian by default; ternary ablations included).
Unless noted, we use $D{=}10{,}000$ and $d{=}4{,}096$.

\medskip
\noindent\textbf{Our model (\DecoHD{}).}
We train directly in a decomposed HDC parameterization with $N\in\{1,2,3\}$ layers and per-layer channel counts $L_i\in\{1,\dots,5\}$ (grid over all nonempty tuples).
Each channel is generated from a low-dimensional latent (default $d{=}4096$) through a frozen random projection into $\mathbb{R}^{D}$.
Given an encoded input $h$, stacked binding ($\otimes$) across layers yields $M{=}\prod_i L_i$ path hypervectors that are aggregated by a lightweight class bundling head $W\in\mathbb{R}^{C\times M}$ via $\oplus$.
Logits are dot products with $h$; we train with cross-entropy.
Optimization uses AdamW (learning rate $1\!\times\!10^{-3}$, weight decay $5\!\times\!10^{-5}$), batch size $1024$ with microbatches of $128$, for $1000$ epochs.
$W$ is initialized uniformly to $1/M$; the input and per-layer projections remain fixed.
Inference follows the same bind $\rightarrow$ bundle $\rightarrow$ dot pipeline.

\medskip
\noindent\textbf{Baselines.}
We compare against (i) a conventional HDC prototype table constructed by class-wise summation and (ii) \emph{OnlineHD} with iterative refinement for $200$ epochs (learning rate $0.1$).
For feature-axis reduction, we include \emph{SparseHD} under matched memory budgets.
All baselines share the same encoder $\phi$ and preprocessing and are evaluated under identical precision settings.

\medskip
\noindent\textbf{Target memory budgets.}
Results are reported under budgets $\le m\in(0,1]$ and denoted in parentheses as $(\le m)$, where $m$ is the model size relative to a conventional HDC table with $C\times D$ parameters (at $b$ bits each). \DecoHD{} replaces this table with a lightweight bundling head $W\!\in\!\mathbb{R}^{C\times M}$ and a bank of channel latents totaling $L_{\mathrm{tot}}=\sum_i L_i$ of size $d$, yielding the approximate normalized footprint
$m \;\approx\; \frac{C M + L_{\mathrm{tot}} D}{C D},$
with $M=\prod_i L_i$. Selecting $(N,\{L_i\},D)$ to satisfy $m$ enforces the target budget.

\subsection{Performance of \DecoHD{}}

\autoref{fig:decohd_perf} compares \DecoHD{} to the strong non-reduced HDC baseline, OnlineHD, across numeric precisions and hypervector dimensionalities $D$ under target memory budgets $m$ (values shown in parentheses). Despite operating with substantially fewer trainable parameters, \DecoHD{} tracks OnlineHD closely across datasets and settings. At $D{=}10\mathrm{K}$, the average accuracy gaps are small—approximately $0.15\%$ at $m{\le}0.7$ and $0.1\%$ at $m{\le}0.5$—with a worst case of about $5.7\%$ even at the tight $m{\le}0.5$ budget. When both precision and dimensionality are reduced, the gap grows modestly as expected (e.g., roughly $0.7\%$ at 16-bit and $D{=}1\mathrm{K}$ for $m{\le}0.5$), yet remains minor in absolute terms. Two trends recur across settings. First, lower numeric precision slightly increases the gap to OnlineHD but does not alter the qualitative ranking, indicating that the decomposition does not introduce precision-specific failure modes. Second, lowering $D$ reduces the usual concentration-of-measure benefits of HDC; nevertheless, \DecoHD{} remains competitive, suggesting that composing class prototypes from a shared channel bank preserves much of the structure that OnlineHD leverages at full size. Overall, across budgets, precisions, and dimensionalities, \DecoHD{} delivers accuracy within a tight margin of OnlineHD while meeting strict memory targets, demonstrating that training directly in the decomposed parameterization is an effective route to compress HDC models with negligible loss.

\subsection{Comparison to feature-axis reduction}

\autoref{fig:decohd_perf_comp} contrasts \DecoHD{} with SparseHD, a state-of-the-art feature-axis reduction method that compresses by shrinking $D$. Under matched budgets, \DecoHD{} outperforms SparseHD in all cases for $m{\le}0.7$ and in six of eight cases for $m{\le}0.5$. The difference reflects where compression is applied. Feature-axis reduction weakens high-dimensional orthogonality and the concentration properties central to HDC’s separability and robustness; as $D$ is reduced, prototype quality and similarity estimates degrade, and the classifier becomes more sensitive to noise and incidental correlations. \DecoHD{}, by contrast, preserves the full ambient dimensionality and instead compresses along the class axis via decomposition, sharing a compact set of expressive channels across classes using binding and bundling. The class head $W$ then allocates these shared channels adaptively. This strategy is more resilient at tight budgets because it amortizes parameters over classes while retaining the geometric advantages of a large $D$.

\subsection{Robustness of \DecoHD{}}

\autoref{fig:decohd_robust} evaluates tolerance to random bit-flip noise injected into 32-bit floating-point representations. At a fixed $D{=}10\mathrm{K}$, \DecoHD{} maintains higher accuracy than OnlineHD as the flip probability increases, indicating that the decomposition does not merely match clean accuracy but also confers stability under perturbation. The effect can be understood through two complementary mechanisms that operate within the same encoder and hyperspace. First, bundling averages over multiple bound channels, so perturbations that affect individual channels tend to be attenuated when aggregated, reducing the variance of the effective class representation encountered at inference. Second, binding behaves like a quasi-orthogonalizing transformation; independent bit flips in one channel yield perturbations that are largely decorrelated from those in other channels, limiting coherent error accumulation. Since both methods share the same input encoding and dimensionality, the robustness margin is attributable to \DecoHD{}'s representation strategy rather than differences in $D$ or preprocessing, which is particularly relevant for near-memory or in-sensor deployments where soft errors and read-disturb effects are non-negligible.

\subsection{Impact of latent dimensionality}

\autoref{fig:decohd_ab_latent} studies latent sizes $d\in\{256,1024,4096\}$ at $D\in\{1{,}000,10{,}000\}$ using a two-layer $\lfloor\sqrt{C}\rfloor\times\lfloor\sqrt{C}\rfloor$ configuration. Accuracy generally saturates by $d{=}4096$ across datasets, indicating that channel latents need not match the ambient dimension to achieve strong performance. Tasks with fewer classes, such as PAGE and PAMAP2, reach their accuracy plateau as early as $d{=}256$, reflecting lower intrinsic class complexity and correspondingly lighter demands on channel expressivity. Because the latent bank scales with $L_{\mathrm{tot}} d$, choosing $d\!\ll\!D$ yields substantial parameter savings: in our settings, up to $97.44\%$ fewer trainable parameters compared to directly learning a full $C{\times}D$ prototype table, with minimal or no loss once the dataset-specific plateau is reached. A practical guideline is to select the smallest $d$ on the observed saturation plateau for the target dataset; combined with a modest $M$, this typically meets the memory target $m$ while preserving accuracy.

\subsection{Impact of the number of layers}

\autoref{fig:decohd_ab_layers} varies the number of layers $N\in\{1,2,3\}$ (with up to five channels per layer) across memory budgets for $d\in\{256,4096\}$. The results reveal a clear interaction between latent capacity and architectural depth. When $d$ is large ($4096$), fewer layers often achieve the best accuracy at a fixed budget, because channels are already expressive and additional binding primarily increases $M$—and thus the size of $W$—without commensurate gains. When $d$ is small ($256$), increasing $N$ is beneficial: deeper binding expands the combinatorial basis of path hypervectors and improves class separability at the same overall budget, effectively trading a modest increase in head size for a disproportionately large increase in representational diversity. From a budget-allocation standpoint, recall that $M=\prod_i L_i$ grows multiplicatively with $N$; increasing layers shifts budget from the latent bank ($L_{\mathrm{tot}} d$) toward the class head ($C M$). At small $d$, the marginal utility of enlarging $M$ is high, whereas at large $d$ it is lower because channels already encode rich structure. In practice, given a target $m$, it is effective to first pick the smallest $d$ that lies on the latent-dimension plateau (as identified in \autoref{fig:decohd_ab_latent}); if this $d$ is small, favor $N\ge 2$ with moderate per-layer channel counts to grow $M$, whereas if $d$ is large, prefer $N\in\{1,2\}$ with slightly larger per-layer channel counts and allocate remaining budget to the class head. This recipe consistently stays within budget while preserving or improving accuracy.

\subsection{Hardware Efficiency and Trade-offs}

\begin{table}[t]
\centering
\caption{System-level efficiency of 1-layer \DecoHD{} (ASIC) vs. a conventional HDC baseline executed on CPU/GPU/ASIC (ISOLET; $D{=}10{,}000$, $C{=}26$, $L_i{=}10^{1/N}$).}
\label{tab:decohd_hw_efficiency}
\resizebox{\columnwidth}{!}{%
\begin{tabular}{lccc}
\toprule
\textbf{Platform} & \textbf{Energy eff. ($\times$)}~$\uparrow$ & \textbf{Speedup ($\times$)}~$\uparrow$ & \textbf{Memory usage ($\times$)}~$\downarrow$ \\
\midrule
CPU (Ryzen 9 9950X) & 277.40 & 34.87 & 0.38 \\
GPU (RTX 4090)      & 13.53  & 3.66  & 0.38 \\
ASIC                & 2.02   & 2.42  & 0.38 \\
\bottomrule
\end{tabular}
}
\end{table}

\begin{table}[t]
\centering
\caption{Depth trade-offs of \DecoHD{} (ASIC) vs. a conventional HDC (ASIC) on ISOLET ($D{=}10{,}000$, $C{=}26$, $L_i{=}10^{1/N}$).}
\label{tab:decohd_hw_efficiency2}
\resizebox{\columnwidth}{!}{%
\begin{tabular}{lccc}
\toprule
\textbf{\# Layers} & \textbf{Speedup ($\times$)}~$\uparrow$ & \textbf{Memory usage ($\times$)}~$\downarrow$ & \textbf{Accuracy drop (\%)}~$\downarrow$ \\
\midrule
1-layer & 2.42 & 0.38 & 1.01 \\
2-layer & 0.94 & 0.12 & 9.81 \\
3-layer & 0.89 & 0.08 & 28.7 \\
\bottomrule
\end{tabular}
}
\end{table}

\autoref{tab:decohd_hw_efficiency} shows that mapping \DecoHD{} to an ASIC yields up to $277\times$ higher energy efficiency and strong speedups over CPU/GPU baselines, while remaining both faster and more energy-efficient than a conventional HDC ASIC, at only $0.38\times$ memory. As \autoref{tab:decohd_hw_efficiency2} indicates, depth induces a clear trade-off: a shallow (1-layer) design maximizes throughput with minimal accuracy loss; increasing depth further reduces memory but increases test time per-sample compute (reducing speed) and increases per-channel interference (reducing accuracy). Therefore, our findings suggest selecting the shallowest factorization that meets the memory budget to preserve throughput and accuracy; deeper configurations are justified only when memory is the primary system bottleneck.

%% file: Sections/5_Conclusions.tex
We presented \DecoHD{}, a gradient-based class-axis decomposition for HDC that enables memory-lean inference under extreme resource constraints. Across benchmarks, it closely matches a strong non-reduced HDC baseline (average accuracy gap $\approx$0.1–0.15\%), improves robustness to random bit-flip noise, and reduces trainable parameters by up to $\sim$97\% at saturation. In hardware, an ASIC realization achieves up to $277\times$ higher energy efficiency at $0.38\times$ memory, with consistent speedups over CPU/GPU and a conventional HDC ASIC. By preserving full hypervector dimensionality while aggressively reducing stored prototypes, \DecoHD{} is well suited for TinyML and edge deployments in memory-first and near-memory systems. Depth reduces memory but increases test-time computation; thus, the shallowest factorization that meets the target memory budget is preferred to preserve throughput and accuracy.